\documentclass{nature}

\usepackage{amssymb} 
\usepackage[implicit=false]{hyperref}
\usepackage{bbm}
\usepackage{mathrsfs}
\usepackage{amsfonts}
\usepackage{graphicx}
\usepackage{cite}
\usepackage[linesnumbered,ruled,lined]{algorithm2e}
\usepackage{array}
\usepackage{subfigure}
\usepackage{multirow}
\usepackage{color}
\usepackage{amsmath}
\usepackage{bm}
\usepackage{upgreek}
\usepackage{float}

\usepackage{cite}
\usepackage{authblk}
\usepackage{stfloats}
\usepackage{multirow}
\usepackage{bbding}
\usepackage{url}

\title{Image-free multi-character recognition}

\author[]{Huayi Wang,$^1$ Chunli Zhu,$^1$ Liheng Bian,$^{1,*}$ 
}

\begin{document}

\maketitle

\begin{affiliations}
 \item School of Information and Electronics \& Advanced Research Institute of Multidisciplinary Science, Beijing Institute of Technology, Beijing 100081 \\
 $^*$ bian@bit.edu.cn
\end{affiliations}


\begin{abstract}
The recently developed image-free sensing technique maintains the advantages of both the light hardware and software, which has been applied in simple target classification and motion tracking. In practical applications, however, there usually exist multiple targets in the field of view, where existing trials fail to produce multi-semantic information. In this letter, we report a novel image-free sensing technique to tackle the multi-target recognition challenge for the first time. Different from the convolutional layer stack of image-free single-pixel networks, the reported CRNN network utilities the bidirectional LSTM architecture to predict the distribution of multiple characters simultaneously. The framework enables to capture the long-range dependencies, providing a high recognition accuracy of multiple characters. We demonstrated the technique's effectiveness in license plate detection, which achieved 87.60$\%$ recognition accuracy at a 5$\%$ sampling rate with a higher than 100 FPS refresh rate. 
\end{abstract}

\newpage

High-efficiency perception under resource-constrained platforms has drawn much attention, especially in the domain of target classification \cite{ota2018ghost, Cao:21}, detection \cite{zhou2021non}, and tracking \cite{deng2020image}. However, in the common imaging-based perception applications, the non-target areas for semantic recognition increase the burden of both communication hardware resources and reconstruction algorithms. Therefore, the conventional "imaging first-perceive later" model is not the best choice for machine intelligence in most cases. A single-pixel detector (SPD) is an effective light coupling device as one of the emerging perception sensors, which could effectively improve the data compression rate and reduce the storage of unnecessary redundant information. In combination with a spatial light modulator (SLM), SPD could modulate and compress the scene during the acquisition process to finish image-free sensing tasks \cite{13bian2018experimental, edgar2019principles}.

\begin{figure}[t]
	\centering\includegraphics[width=0.7\linewidth]{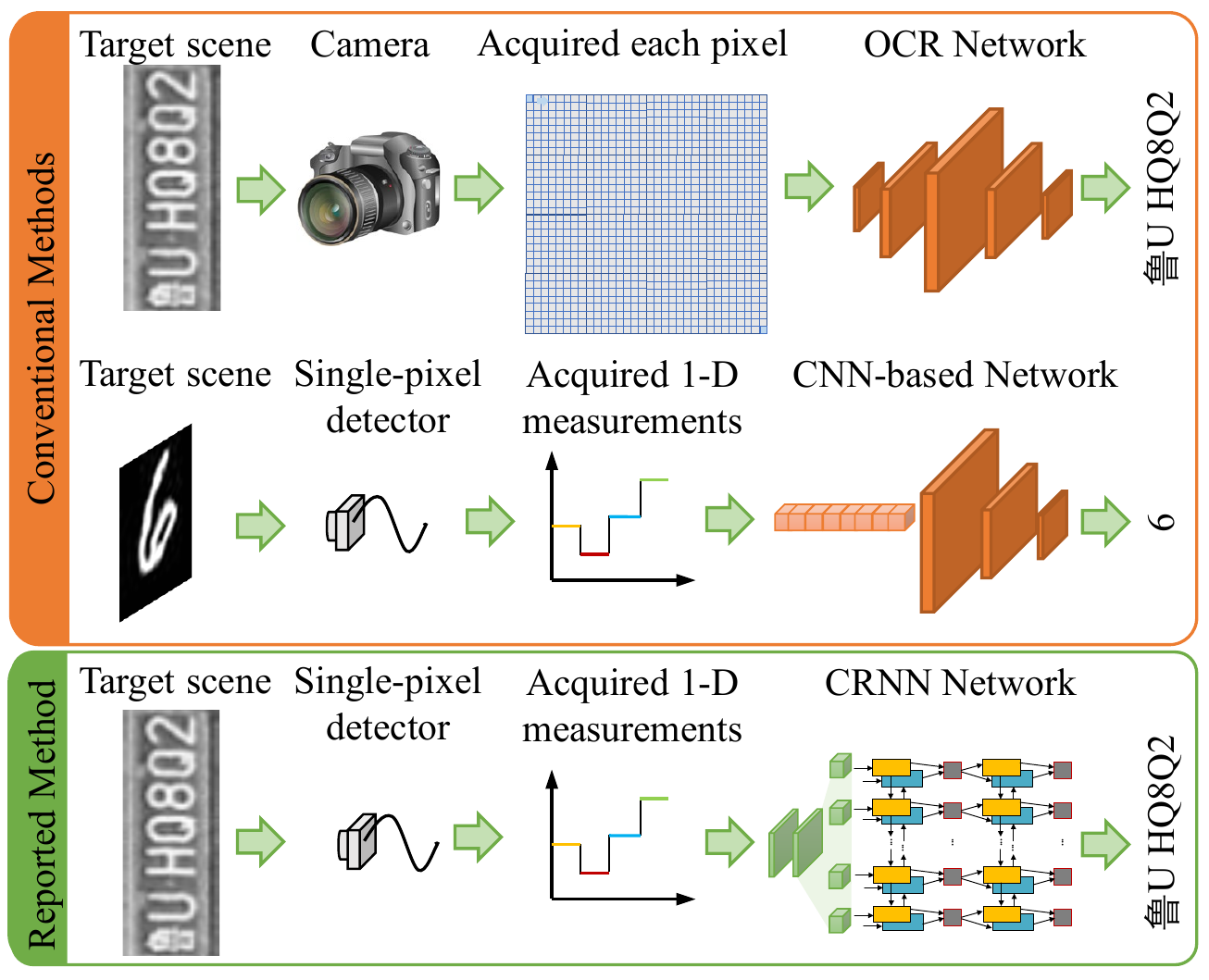}
	\caption{Comparison between conventional optical character (OCR) recognition, conventional single-pixel classification, and the reported method. Multi-target information can be inferred directly from a small number of single-pixel measurements using the CRNN.}
	\label{fig:setup}
\end{figure}

Driven by the deep learning enabled analytical power \cite{long2021scene, bartz2018see}, studies on single-pixel image-free methods have witnessed an incremental increase in recent years. For the relatively simple single-semantic information classification task, Lohit \emph{et al.} \cite{lohit2016direct} introduced convolutional neural networks (CNNs) to extract the differential nonlinear features from the compressed measurements for the next-step target classification. Additionally, an EfficientNet-based decoding net utilized in H. Fu \emph{et al.}'s work \cite{fu2020single}, targeted at directly handwritten characters' recognition by optimizing the modulation patterns in the encoding module. Moreover, Z. Zhang \emph{et al.} \cite{zhang2020image} reported an online fast-moving digits classification task with fully connected layer. However, the previous works (summarized as the lower part of the tangerine box in Fig. \ref{fig:setup}) lack the capability in multi-semantic information recognition when considering the accuracy and speed.

\begin{figure}[h]
	\centering\includegraphics[width=0.7\linewidth]{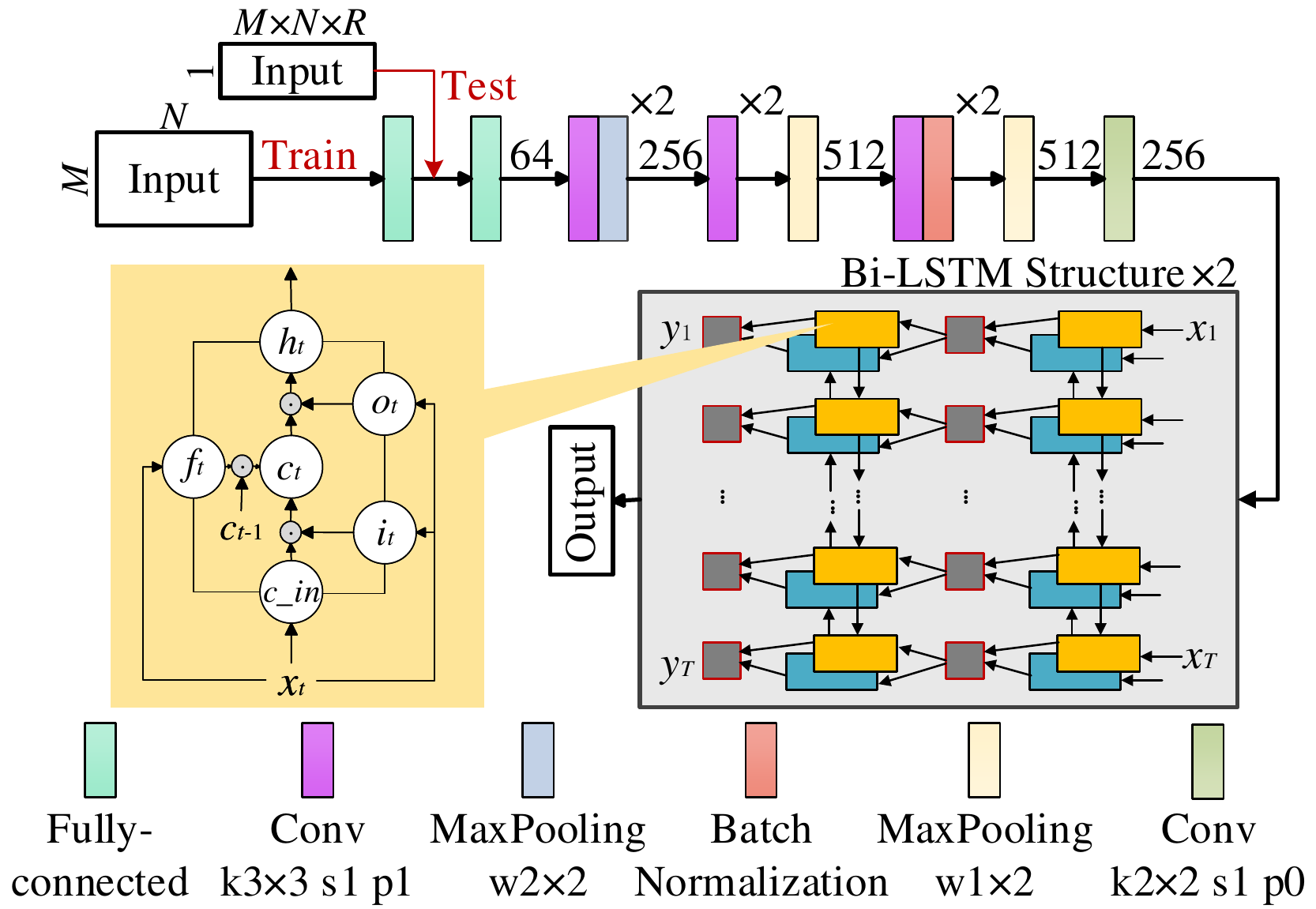}
	\caption{The overall CRNN structure, where $k$, $s$ and $p$ represent the kernel size, stride and padding size, respectively. The deep Bi-LSTM consists of multiple forward (top to bottom) and backward (bottom to top) LSTMs, where the yellow block illustrates the specific formulas and network update details of each hidden layer.}

	\label{fig:network}
\end{figure}

Here, we report a multi-target recognition oriented image-free sensing for the first time. The utilized CRNN based architecture (as in the green box of Fig. \ref{fig:setup}) could perceive multiple targets directly. In Fig. \ref{fig:network}, the overall architecture of the CRNN (coupled CNN and LSTM) is demonstrated, in which the first fully-connected layer is on behalf of the SPD modulation, therefore integrating the sensing process as a whole. The described end-to-end multi-target recognition network is open-sourced on \url{https://bianlab.github.io}.

For further details, the CNN module, which is used to extract the sequence feature representation of the input image, is composed of two fully connected layers, multiple convolutional layers, and pooling layers. Here, the first and second fully connected layers are designed for the functionality of training the optimized modulation without bias and performing the feature inference, respectively. It is worth noticing that the training process starts from the first fully connected layer, while the test process starts from the second one. Therefore, the input dimensions of those two layers differ, where $M\times N$ represents the image size and $R$ is the sampling rate (SR). The SR is defined as the ratio of measurements number to pixels number. With the followed convolution and pooling operations, a total of 256 one-dimensional feature vector $\{x_{1},...,x_{T}\}$ are extracted from the feature map generated by the convolutional layers.

In terms of the LSTM module, the output sequences $\{x_{1},...,x_{T}\}$ of CNN are the input of the bidirectional-LSTM (Bi-LSTM). As a representative kind of RNN, Bi-LSTM \cite{greff2016lstm} is adopted here for capturing the two-way semantic dependence to model the image-based sequences, which is different from semantic analysis or time series that only uses the past information. Formulated by the stacked Bi-LSTMs, the deep Bi-LSTM enables higher level abstractions that make the directly multi-target character recognition feasible. In each Bi-LSTM, the forward and backward LSTMs make the full use of the specific image-based sequences from the past and future information. The final output of the network is the judgement features of $\{y_{1},...,y_{T}\}$.

We use $X = \{I_{i}, \mathbf{I}_{i}\}$ to represent the training dataset, where $I_{i}$ is the training image and $\mathbf{I}_{i}$ is the ground truth label sequence. The goal is to minimize the negative log likelihood of the conditional probability of ground truth \cite{graves2006connectionist}, which reads,

\begin{equation}\label{eq:1}
O=-\sum_{I_{i}, \mathbf{I}_{i} \in x} \log p\left(\mathbf{I}_{i} \mid y_{i}\right),
\end{equation}

\noindent where $y_{i}$ is the sequence generated by the $\mathbf{I}_{i}$ 's cyclic and convolutional layers. Eq.\ref{eq:1} calculate the cost directly with the image and its ground truth label sequence, thus eliminating the manually marking process of training images' individual components.

The units in each LSTM are the yellow (forward) and blue (backward) rectangular blocks in Fig. \ref{fig:network}. The forward and backward LSTMs have the same calculation process. Each unit reads,


\begin{equation}\label{eq:2}
\hat{c}(t)=f(W \cdot x_{t}+V \cdot y_{t-1}),
\end{equation}

\noindent where $x_{t}$ and $y_{t}$ are the input and output sequence of time slice $t$; $y_{t-1}$ is the output state of the previous moment; $\hat{c}(t)$ is the instantaneous state of the cell at $t$; $W$ and $V$ are the respective weights of the corresponding sequences.

The calculation of $\hat{c}(t)$ in Eq. \ref{eq:2} does not consider the state of previous step $c(t-1)$, therefore the gradient or error will be exponentially attenuated or amplified with forward propagation. So $c(t)$, the state of a cell at time slice $t$, is calculated as,

\begin{equation}\label{eq:3}
c(t)=c(t-1) \cdot \hat{c}(t),
\end{equation}

Combining Eq. \ref{eq:2} and Eq. \ref{eq:3}, the output of an unit can be obtained by $y_{t}=f(c(t))$, where $f(\cdot)$ is the activation function. Each input will go through the forget gate $f_{t}$, information enhancement gate $i_{t}$ and output gate $o_{t}$, as shown in the yellow box of Fig. \ref{fig:network}. Each gate function get the final $y_{t}$ sequence after the above calculations, which is the desired multi-semantic vector.

In addition, stochastic gradient descent (SGD) \cite{bottou2012stochastic} is used in the CRNN network. In the transcription layer, the loss is backpropagated via the forward-backward algorithm; in the cyclic layer, backpropagation through time (BPTT) is introduced to calculate the error differential \cite{shi2016end}. ADADELTA \cite{zeiler2012adadelta} is used to automatically calculate the learning rate for each dimension, which eliminates manual setting of the learning rate for the momentum method \cite{qian1999momentum}.

Hereafter, we will discuss the modulation process via the first fully connected layer, in which the original weight $W_{o}$ is generally between -1 and 1. One simple attempt for reducing the amount of modulation is to restrict the weights to be positive values. Assuming that the adjusted weight is $W_{a}$, the modulation pattern is close to the weight, namely, $W_{o} \approx W_{a}$. Therefore, we artificially stipulate the weight $W_{a}$, if $W_{o} \in [0,1]$, $W_{a}$ remain unchanged, where no loss is increased; and if $W_{o} \in [-1,0]$, $W_a$ is set as equal to $-W$, where the loss is greater than $0$. Therefore, the best gray scale weight is equivalent to solving the following optimization problem \cite{13bian2018experimental}:

\begin{equation}\label{eq:4}
W_{a}=\underset{W_{a}}{\arg \min }\left\|W_{o}-W_{a}\right\|^{2},
\end{equation}

Specifically, the objective function could be rewritten as  $F\left(W_{a}\right)=W_{o}^{T} W_{o}-2 W_{o}^{T} W_{a}+W_{a}^{T} W_{a}$. 
The gray-scale weight $W_a$ is first calculated via Eq. \ref{eq:4}, and then the original unconstrained weights are replaced with previous constrained weights.

For further improvement, we add a training stage to keep the weight of coding pattern fixed and update the rest layers \cite{fu2020single}. When convergence, the obtained optimized gray-scale weight would be adjusted to the ideal light modulation pattern. Then, a 1-D coupling measurement value could be obtained via SPD with the adjusted pattern. After the measured values are put into the first fully connected layer, the network trained in the second stage directly outputs the recognition results.

\begin{table}
	\centering
	\caption{\bf Network Usability Experiments}
	\begin{tabular}{lllll}
		\hline
		Network  & Imaging & Image-free \\
		\hline
		MTCNN \cite{wang2019light}     & \Checkmark (96.91\%)       & \XSolidBrush   &     \\
		\hline
		
		MTCNN +  STN \cite{jaderberg2015spatial} & \Checkmark  (97.30\%)     & \XSolidBrush       \\
		\hline
		OCR-net \cite{silva2018license}      & \XSolidBrush     & \XSolidBrush      \\
		\hline
		LPR-net \cite{zherzdev2018lprnet}      & \Checkmark  (95.00\%)    & \XSolidBrush          \\
		\hline
		RNN  \cite{xu2018towards}     & \Checkmark   (96.13\%)     & \XSolidBrush    \\
		\hline
		\bf{CRNN \cite{shi2016end} (Ours) }   & \bf{\Checkmark (98.25\%)  }      & \bf{\Checkmark (92.73\%) }      \\
		\hline
	\end{tabular}
	\label{table:network}
\end{table}

We complete the verification of the CRNN on the Chinese City Parking Dataset (CCPD) dataset with license label \cite{xu2018towards}. The data volume of the CCPD is about 260,000, with unified resolution 720 (width) $\times$ 1160 (height) $\times$ 3 (channels), that under different light (light and dark), angles (different inclination angles), and weather conditions (rain, snow, fog, etc.). A total of 2,600 license plate images were tested. For convenience, we cropped images uniformly with the size $M=96$, $N=32$, to generate new images with license plate characters only. In each license, there are 7 target characters, the first character is the province and the last 6 characters are a mixture of numbers and letters without repetition. During training, the input is the cropped image, and output features of the fully connected layer adapt according to the number of modulation patterns, which $M\times N \times R $ is set as 100, 150, 200, and 250. Referring to \cite{xie2018new,xu2018towards,shi2016end}, the specified recognition accuracy rate is the ratio between the number of the correctly recognized samples and the total samples; for a single sample, all characters in each sample are correctly recognized before it is counted as the number of correct recognition.

We conducted simulations to validate the feasibility of the CRNN in multi-character recognition. Several existing neural networks were applied for imaging and image-free perception, as shown in Table. \ref{table:network}. A cross mark indicates the inefficiency of the network training, either with gradient disappears/explodes, or the accuracy fluctuating at approximately 20\%. The value in the brackets following the checkmark is the overall recognition accuracy rate after training for the same epochs. The table shows that the CRNN network demonstrates a 98.25\% accuracy with imaging, while the revised image-free network still exhibits a high accuracy value of 92.73\% after transplanting the conventional network to a single-pixel network. However, the conventional CNN-based single-pixel imaging network structures (such as MTCNN) and OCR networks (such as LPR-net) cannot tackle the targeted single-pixel multi-character task.

\begin{table}
	\centering
	\caption{\bf Comparison between Various SR and Patterns on the Same License Plate Test Set}
	\begin{tabular}{lllll}
		\hline
		\multirow{2}{*}{Modulation Mode} & \multicolumn{4}{l}{Accuracy (\%) with various SR} \\
		& 0.03       & 0.05       & 0.07       & 0.09       \\
		\hline
		\{-1,+1\} \cite{higham2018deep}   & 40.52      & 46.98      & 52.16      & 58.52      \\
		\hline
		Random{[}0,+1{]} \cite{zhang2020image}     & 62.63      & 76.53      & 80.72      & 89.28      \\
		\hline
		\bf{Optimized{[}0,+1{]}} (Ours)   & \bf{82.32}      & \bf{87.73}      & \bf{90.283}     & \bf{92.73}      \\
		
		\hline
	\end{tabular}
	\label{table:pattern}
\end{table}

For training details, the sampling rates of $0.03$ and $0.05$ for the first training stage require $100$ epochs, while the sampling rates of $0.07$ and $0.09$ require $200$ epochs. After the first stage, the modulation is fixed, and the training procedure is terminated if the performance of the validation set has not improved with $100$ epochs. The whole training process takes 4-5 hours. We implement the network within the PyTorch framework. The computer is equipped with an Intel i7-9700K processor (3.6GHz) and 64GB RAM and an NVidia RTX 3090 graphic card.

As to the selection of most suitable modulation patterns, we first note that Higham \emph{et al.} \cite{higham2018deep} reported a two-stage training method to binarize the modulation pattern, thereby normalizing the pattern to approximately \{-1,+1\}. However, that work requires two complementary patterns to achieve the negative modulation, which doubles the modulation and acquisition time. Therefore, we refer to the article \cite{fu2020single} to restrict the pattern weight. Due to the complexity of multi-semantic tasks, we finally got optimized gray-scale modulation patterns for the reported multi-target recognition network CRNN. As in Table. \ref{table:pattern}, we use inverse modulation pattern \cite{higham2018deep}, gray-scale modulation pattern \cite{zhang2020image}, and the optimized pattern for comparison. Results showed that the binary pattern performed not well due to the information loss, in which the accuracy rate did not exceed 60\%. So we choose the optimized gray-scale pattern for the balanced sampling rate and accuracy.

\begin{figure}[t]
	\centering\includegraphics[width=0.7\linewidth]{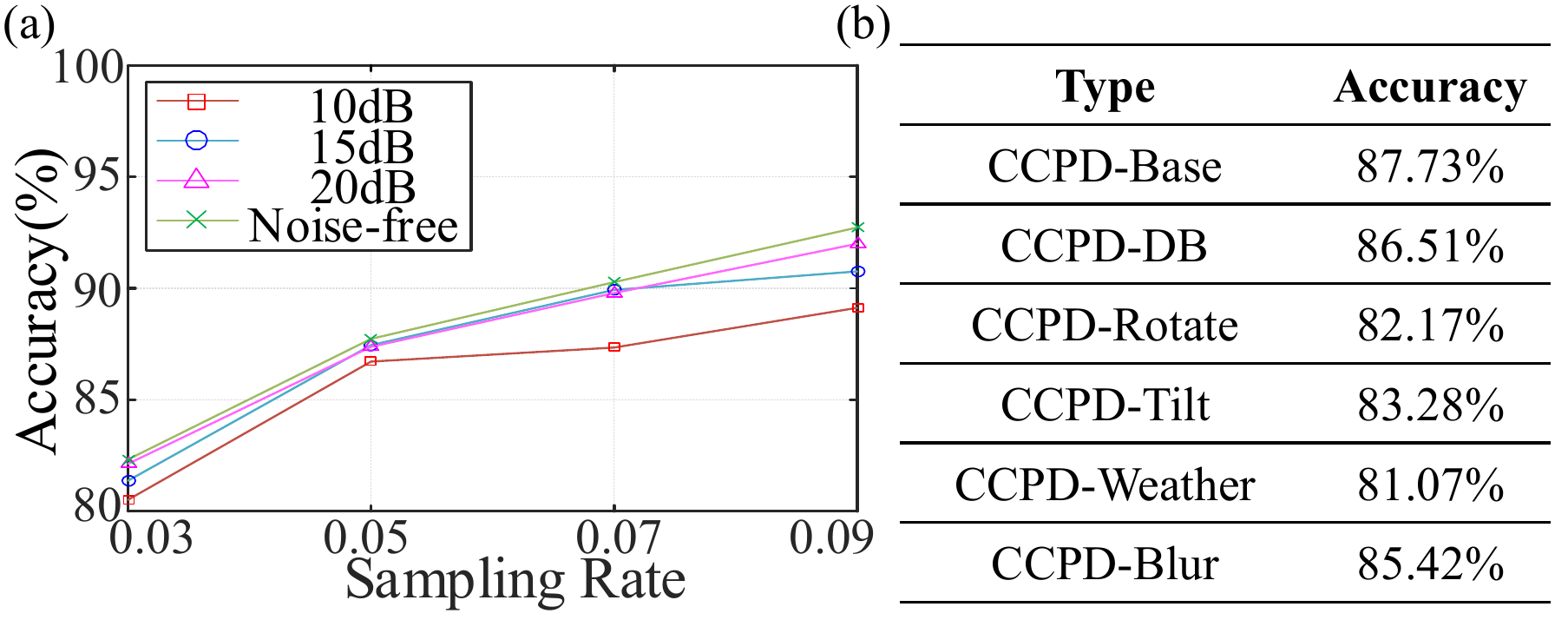}
	\caption{Accuracy of reported method in various situations. (a) Under various sampling rates and levels of noise. (b) Comparison of various data sets under 5\% sampling rate, where -Base is the standard data selected at random, -DB is the license plate under dark or strong light, -Rotate and -Tilt are the license plates rotated and tilted at a small angle respectively, -Weather is the license plate under rain/snow/fog, and -Blur is the blurred license plate taken.}
	\label{fig:compare}
\end{figure}

\begin{figure}[t!]
	\centering\includegraphics[width=0.7\linewidth]{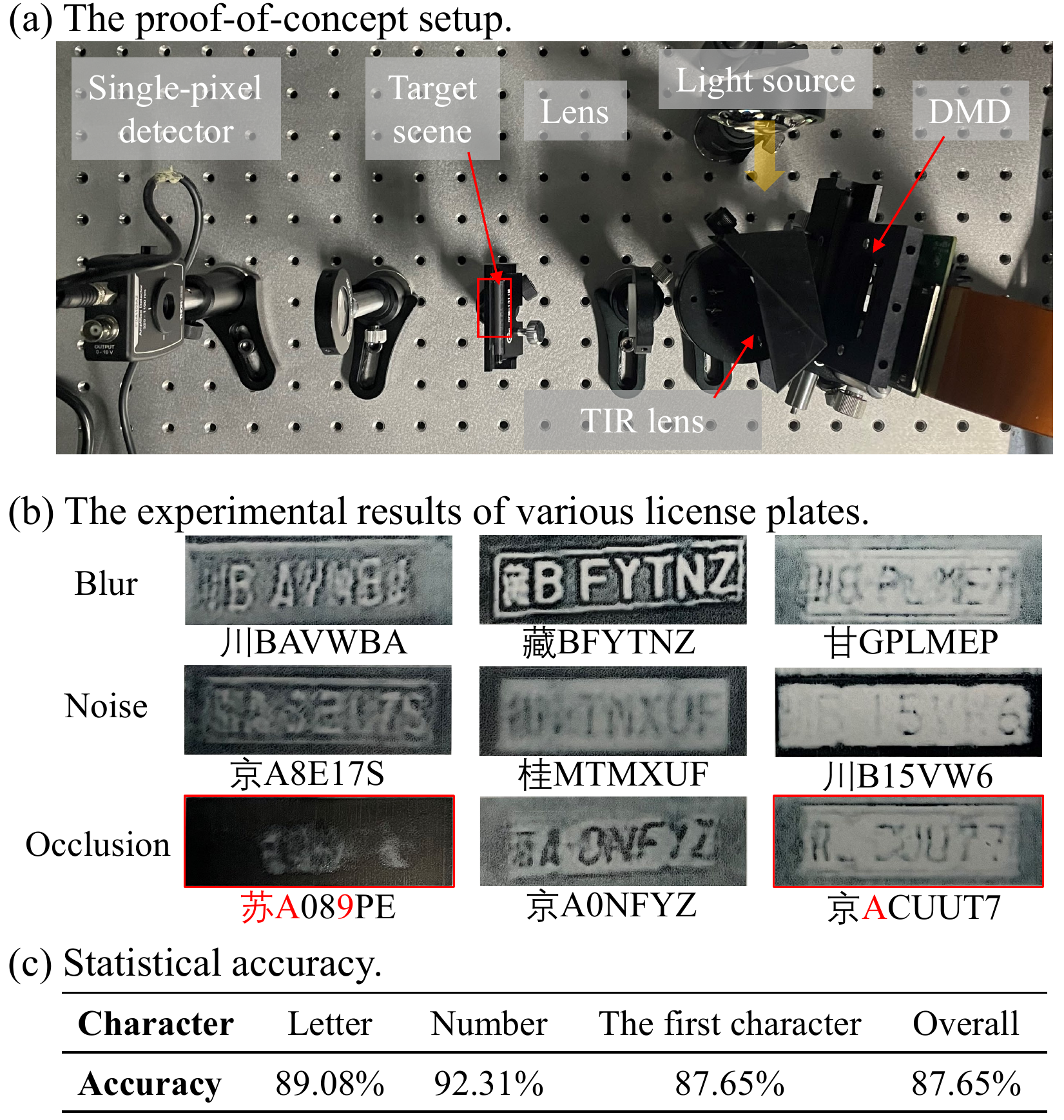}
	\caption{(a) The proof-of-concept setup for single-pixel multi-character recognition. (b) The experimental results of reported CRNN for various license plates at 5\% sampling rate (the red box is the scene with the wrong recognition). (c) Statistical accuracy of a single character and the overall accuracy. }
	\label{fig:experiment}
\end{figure}

We then discuss the impact of sampling rates (0.03–0.09) on classification accuracy. Fig. \ref{fig:compare} (a) shows the recognition performance simulated on the CCPD test dataset. Results showed that when the sampling rate increases from 0.03 to 0.09, the overall recognition accuracy of the license plate dataset increases from 82.32\% to 92.73\%. However, accuracy would not significantly be improved with the sampling rate higher than 0.05. Therefore, the sampling rate is chosen as 0.05 (corresponding to 150 measurements) in the following simulations and experiments.


To verify the robustness of the network to noise, we also added different degrees of Gaussian noise to the dataset. Noise with an SNR of 10 dB to 20 dB is added to the one-dimensional measurement. The result in Fig. \ref{fig:compare} (a) shows that the reported method could maintain higher recognition accuracy when there exists measurement noise, even at 10dB SNR and 5\% sampling rate, there is an accuracy of 86.72\%. In addition, we have selected some special environment datasets to verify the SPD and the CRNN multi-target recognition network's performance under low-light and harsh conditions (including dim light, tilt, bad weather, etc.). The specific experimental results are shown in Fig.\ref{fig:compare} (b), in which the accuracy rate can still obtain better results (87.73\%) when the characters are blurred or slanted.


We built a proof-of-concept setup as shown in Fig. \ref{fig:experiment} (a). The method of active lighting was adopted. The structured illumination was generated by using a digital micromirror device (DMD) (V-9501, Vialux, 22KHz) and a 10-watt white LED (488-SF-CDRH, SAPPHIRE) as the light source. A single-pixel detector (PDA100A2, Thorlabs) was used to collect the light intensity through the target scene. The collected 1-D measurements were digitalized using a data acquisition card (PCIE8514, 10MHz) and fed to the trained CRNN after the first fully connected layer on the computer. The license plates printed on the film were used as the target scene, and the light was transmitted through the target. We select 150 gray-scale patterns (approximately $96 \times 32 \times 0.05$) under 5\% sampling rate for each sample and put them into DMD. The dithering strategy proposed by Z. Zhang \emph{et al.} \cite{zhang2020image} can be used to further reduce the sampling rate.

The multi-character recognition results are shown in Fig. \ref{fig:experiment} (b), (c). The overall modulation time of each scene by the spatial light is $0.0047s$. The average recognition time of a single license plate is $0.0027s$, which is better than performing scene segmentation first and then recognizing single character ($0.005s \times 7$) \cite{fu2020single} as well as the time to directly reconstruct the scene ($0.05s$) \cite{higham2018deep}. Fig. \ref{fig:experiment}(b) shows the recognition results corresponding to several real scenes printed on the film, where a few fuzzy characters were mistakenly recognized, especially with the obscured scene. When in the case of blur and noise, the reported method is more stable in recognition and achieves better accuracy than recognition after imaging. Fig. \ref{fig:experiment}(c) shows the statistical accuracy of various characters and the overall accuracy. It can be seen that the recognition accuracy of a single character is significantly improved. 
A possible reason for the relatively lower accuracy of the first character lies in the difficulties of Chinese words recognition.
The combination of pure number and letter license plate data set (such as European license plate data set \cite{gonccalves2016benchmark}) could improve the result through retraining. The overall accuracy rate achieved good results in a short time and under harsh environmental conditions. This provides directions and possibilities for single-pixel detectors to further perceive multiple targets.

In this letter, an image-free multi-target sensing system based on the CRNN network and SPD framework is reported and validated experimentally for the first time. The advantages of the single-pixel multi-character recognition technology lie in the following two aspects. First, an end-to-end recursive network is built to directly perceive multiple targets from coupled measurements, which could reduce the burden of simulation and storage compared with the conventional imaging methods. Second, the gray-scale modulation and the CRNN network are trained together to ensure that the best sensing efficiency could be obtained with the least measurements. The single-pixel multi-target recognition technology has been successfully proved in the recognition task of large-scale license plate data sets. The simulation and experimental results showed that the technology achieved 87\% multi-target recognition accuracy at 100 FPS refresh rate at a sampling rate of 5\%. Due to the strong noise immunity, the device can be combined with a multi-camera system to complete wider tasks in the future, such as infrared light band imaging and multi-character recognition in blurred scenes.

\vspace{5mm}

\bibliographystyle{naturemag}
\bibliography{MSPI}


\begin{addendum}
 \item National Key R\&D Program (Grant No. 2020YFB0505601); National Natural Science Foundation of China (Nos. 61971045, 61827901, 61991451). 
 \item[Competing Interests] The authors declare no competing financial interests.
\end{addendum}


\end{document}